# A survey of datasets for computer vision in agriculture

## A catalog of high-quality RGB image datasets of natural field scenes

Nico Heider[1], Lorenz Gunreben[1], Sebastian Zürner[1] and Martin Schieck[1]

**Abstract:** In agricultural research, there has been a recent surge in the amount of Computer Vision (CV) focused work. But unlike general CV research, large high-quality public datasets are sparsely available. This can be partially attributed to the high variability between different agricultural tasks, crops and environments as well the complexity of data collection, but it is also influenced by the reticence to publish datasets by many authors. This, as well as the lack of a widely used agricultural data repository, are impactful factors that hinder research in applied CV for agriculture as well as the usage of agricultural data in general-purpose CV research. In this survey, we provide a large number of high-quality datasets of images taken on fields. Overall, we find 45 datasets, which are listed in this paper as well as in an online catalog on the project website: https://smartfarm-inglab.github.io/field_dataset_survey/.

## 1 Introduction

Deep learning-based computer vision is delivering state-of-the-art results across various agricultural computer vision (CV) tasks [Ra23; KP18], particularly in precision agriculture, where applications like vision-based precision weeding are becoming increasingly practical [MLS18]. In general-purpose computer vision, the availability of large datasets such as LAION-5B [Sc22], CIFAR [Kr09], and ImageNet [De09] has been instrumental in achieving impressive outcomes in image classification, segmentation, and generation. These datasets have also paved the way for foundation models that excel across diverse tasks [Ra21]. However, in the agricultural sector, data availability is more limited, primarily due to the high cost and complexity of data collection and annotation as well as the relatively small research community that often publishes research without the corresponding datasets. This highlights the need to make the most efficient use of the data that has already been published. To address this challenge, this paper provides a catalog of publicly available RGB image datasets for deep learning in agricultural computer vision.[2] We also aim to encourage the broader computer vision community to use more agricultural image data, as this data inherently exhibits interesting challenges like seasonal data drift and domain shift. Since an up-to-date survey for livestock is available [Bh24], we focus on field scenes and tasks, especially weeding, pest and disease detection. Horticulture, viticulture and aquaculture are beyond our scope. The main contribution of this paper is a list of 45 datasets consisting of high-quality images of pictures taken on fields. While they represent diverse plants, the overall environment the pictures are taken in is visually similar. This enables their usage as training data, combined for self-supervised pre-training of image models [Re22], unsupervised training [Ch24] or image generation [Ru23].

---

[1] Leipzig University, Smart Farming Lab, Grimmaische Straße 12, 04109 Leipzig, heider@wifa.uni-leipzig.de, gunreben@wifa.uni-leipzig.de, zuerner@wifa.uni-leipzig.de, schieck@wifa.uni-leipzig.de

[2] The list is available online at https://smartfarminglab.github.io/field_dataset_survey/.

## 2    Related work

The last review of computer vision datasets for precision agriculture was published more than four years ago [LY20]. Since then, the amount of agricultural computer vision publications has continued to increase exponentially, and the number of available datasets is much larger today (Fig. 1). There are surveys for a variety of other topics in agriculture available, e.g. in weeding [WZW19], crop disease detection [Yu22], plant stresses estimation [He24] or perception systems for agricultural robots [TVG23].[3] They mostly focus on algorithms and techniques and less on the available datasets. Like the authors of [LY20], we focus on datasets representing natural field scenes, i.e. RGB images taken on fields with natural lighting. The main difference to [LY20] is our exclusive focus on field data, where we provide a larger number of datasets, many of which have been published since the release of [LY20]. Another key distinction is that we organize the datasets based on the tasks addressed in the corresponding papers and the plants depicted, making it easier for agricultural computer vision projects to identify suitable data for their specific needs.

The data availability for different modalities like multispectral [Lu20], LIDAR [Ri23] and RGB-D [Ki23] is in general much lower than for RGB images. Three-dimensional reconstruction techniques and their associated datasets, such as [Me24] and [Ma24], are gradually becoming more accessible in the agricultural domain.

Synthetic data is also increasingly prevalent in agricultural CV, e.g. [Me24; Kl24]. Various methods for generating fully or partially synthetic data can help reduce the need for extensive image collection. While a recent comprehensive survey on synthetic data generation is available [Ba24], and another survey specifically on synthetic images exists [MMG24], a survey focused on agricultural CV has yet to be published. Given the specified scope of this work, our survey focuses exclusively on real images. However, we hope that the catalog of datasets presented here will also support the development and testing of synthetic data generation algorithms for agricultural CV.

---

[3] A list of other agricultural computer science surveys is also available at https://smartfarm-inglab.github.io/field_dataset_survey/.

## 3    Materials and methods

We find relevant datasets by manually screening papers tagged 'agriculture' on the pre-print server *arXiv*, screening all 3702 papers – available online at writing time – in the journal *Computers and Electronics in Agriculture* that contain the word 'dataset', as well as searching for 'agriculture' + 'dataset' on Google Scholar and Perplexity[4]. Kaggle and Zenodo were also manually screened for relevant datasets using the search term 'agriculture' and the dataset tag. Other sources were excluded due to limited searchability or lack of relevant content. The sources were selected to encompass a significant portion of published agricultural CV research.

Overall, 136 datasets were in our target domain, of which 45 datasets were picked. We chose the datasets based on several factors: (1) Domain coherence: the data should contain natural field scenes, e.g., plants on fields or pastures taken under natural light; we exclude datasets where the images cannot be cropped to this domain, i.e. low-resolution drone images taken from high attitude.[5] (2) Ground truth data: We only include datasets with a large number of available annotations. (3) Image quality: the images should feature a consistent resolution, low motion blur and adequate lighting. We only include original datasets and exclude datasets that reuse existing images, i.e. web-scraped data. These criteria are somewhat subjective, as applying hard thresholds, such as the number of annotations or image quality measures, would not only be difficult to operationalize – i.e. due to the different types of annotations – but would also exclude datasets we consider relevant. Therefore, the selection process relies on expert judgment to ensure the datasets align with the objectives of this survey.

The large discrepancy between the number of CV publications and the number of datasets can partly be attributed to the repeated usage of popular public datasets, like Global-WheatHeadDetection (GWHD) [Da21], which gets used in multiple publications (e.g. [Ye23; Wa22; Li24a]). But to a larger part, this is caused by the reluctance of many researchers to publish their datasets.

---

[4] https://www.perplexity.ai, accessed 30.10.2024.
[5] MFWD [Ge24], PlantSeedlingClassification [Gi17] and OpenWeedPhenotype [Ma20] violate this rule, containing images taken from plants in soil filled trays under artificial lighting, we still include the datasets because of their large size and high image quality and their ability to be easily cropped to only feature plants and soil.

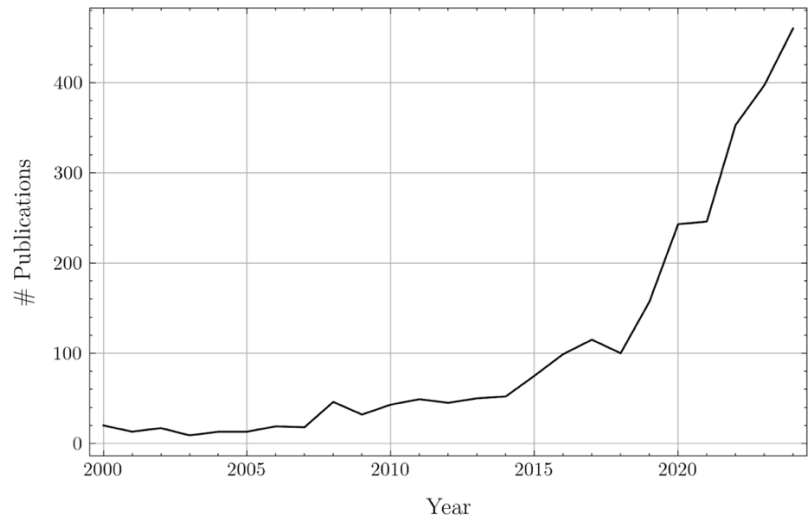

Fig. 1: Number of publications containing the search term 'computer vision' in the journal Computers and Electronics in Agriculture by year. Data retrieved from https://www.sciencedirect.com/journal/computers-and-electronics-in-agriculture, accessed on 30.10.2024

### 3.1    Datasets

The most common task in the featured datasets is weed detection and classification (29), followed by disease and pest detection (9) and seedling or crop detection (6). Other tasks are plant growth stage detection and phenotyping and various detection and counting tasks. The number of images of the datasets varies widely from the small Weed-AI [We24a-m] datasets, which feature few but high-quality images together with a large amount of bounding boxes as annotations, or the SoybeanNet [Li24] dataset, which contains relatively few images but 262,611 manually annotated soybean pods as ground truth data, to large datasets like PhenoBench [We24], where 7 GB of labelled data and additional 50 GB of unlabeled data is available or MFWD [Me24], where more than 900 GB of data is available. Overall, the listed datasets encompass over 2,5 terabytes of image data.

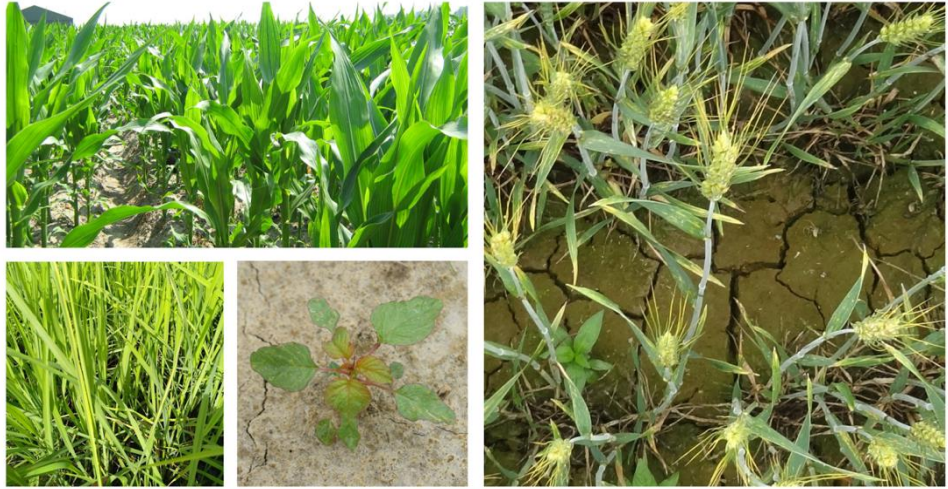

Fig. 2: Example images from included datasets. Top left: [Yo23], right: [Da21], bottom left: [Pe23], bottom middle: [Ch23]

The modality of all datasets is RGB, except for WE3DS [Ki23], which additionally provides depth data (RGB-D). The contents of the datasets can be coarsely categorized by the task they are originally collected for. Other categorizations are available on the project website, where the datasets can be sorted by annotation type, publication year, content, number of images and author. Links to the papers and datasets are also provided.[6]

Plant disease and pest identification:

- CornLeafInfection [Ac20], RiceLeafDiseases [An23], ASDID [BSH22], PaddyDoctor [Pe23], PumkinDiseases [RBH24], MaizeDiseaseSymptoms [Wi18], ibean [Ma20a], SoyNet [RST23], SorghumAphids [Ra24]

Weed detection and classification (listed by crop):

- Cotton: CottonWeedID15 [Ch23], CottonWeedDet3 [RLW23], PalmerAmaranthGrowth [We24m], YOLOWeeds [Da23]

---

- Wheat: NarrabriWheat [We24a], AnnualRyegrassAndTurnipweedInWheat [We24c], WildRadishInWheat [We24i], RadishWheatDataset [We24j], CobbityWheat [We24e], GlobalWheatHeadDetection (GWHD) [Da21]

- Maize: MaizeWeed [Ol22], ImageWeeds (Zea Mays) [Ra23b]

- Chickpeas: AmsinckiaInChickpeas [We24b], NarrabriChickpea [We24f]

- Other or multiple crops: BroadleafWeedsInCommonCouch [We24d], WildCarrot-FlowersInCanola [We24h], ImageWeeds (Brassica) [Ra23b], ImageWeeds (Flax) [Ra23b], ImageWeeds (Field Peas) [Ra23b], ImageWeeds (Lentils) [Ra23b], ImageWeeds (Beta Vulgaris) [Ra23b], NorthernWAWheatbeltBlueLupins [We24g], WE3DS [Ki23], EarlyCropWeed [Es20], RoboWeedMap [TJG22]

- Weeds on grass/pasture areas or standalone weeds: DeepWeeds [Ol19], OpenWeedPhenotype [Ma20], BrownlowHillFireweed [We24k], RumexLeaves [GAN23], ImageWeeds [Ra23b][7]

Seedling and plant classification:

- Seedlings: PlantSeedlingClassification [Gi17], DeepSeedling [Ji19], RyegrassSeedlings [We24l]

- Plants: LucasVision [Yo23], Tobacco Aerial Dataset [Mo23], DPA [RRG20]

Crop segmentation and detection:

- Segmentation: PhenoBench [We24], MFWD [Ge24]

- Growth stage estimation: WeedGrowthState [Te18]

- Detection/Counting: RicePanicles [Ra23a], SoybeanNet [Li24], Sugarcane-PlantCounting [UJ24]

A number of these datasets can be used for multiple supervised tasks, i.e. counting and detection are similar, and phenotyping usually encompasses segmentation. All datasets can be used for unsupervised training or pre-training. Combined to a large field image repository these datasets can also be useful for general-purpose CV research, like domain-shift adaptation between training and test data [Ga16].

---

[7] ImageWeeds is available as comprehensive dataset and as subsets focused on specific crops.

Interoperability is an important factor for the usage of these datasets in practice. While we didn't explicitly choose the datasets based on a common file and annotation structure, we find that many follow the json-based MS-COCO Structure [Li15]. The Weed-AI project from the University of Sydney proposes the WeedCOCO standard, which adds agricultural specific annotations to this structure, like the soil type, crop type and plant growth stages [We24n]. Currently, this is not widely used and none of the datasets besides the one published by this project follow this proposal [We24a-m]. Other datasets like GWHD [Da21] provide a CSV file containing the image names and bounding box coordinates. A popular annotation type for classification tasks is to label the data folders based on the respective classes of their image contents, as seen in the RiceLeafDiseases dataset [An23]. Most datasets contain data that is already split into test and train subsets. Overall, we observe that datasets annotated for specific tasks (e.g., bounding boxes or classification labels) generally use structures that can be adapted with little effort for use alongside other datasets designed for the same task.

## 4    Conclusion

Although the volume of computer vision research within the agricultural domain has kept pace with the rapid growth observed across the broader field of computer vision, the number of high-quality datasets has been somewhat lagging behind. This gap presents a challenge for researchers aiming to build accurate and robust models that can handle the unique complexities of agricultural environments – such as varying weather conditions, changing light, and diverse crop and pest types. High-quality datasets with well-labeled, representative samples are essential for advancing applications like weeding, crop monitoring, pest and disease detection, and phenotyping. The listed datasets serve as great examples of how publicly available data can help to tackle these issues.

### Acknowledgments

This work and the Rubin Feldschwarm® ÖkoSystem project are funded by the German Federal Ministry of Education and Research (BMBF) (grant no. 03RU2U051F, 03RU2U053C).